%% file: BNE_main.tex
\documentclass[final]{cvpr}

\usepackage{times}
\usepackage{epsfig}
\usepackage{graphicx}
\usepackage{amsmath}
\usepackage{amssymb}
\usepackage{comment}
\usepackage{color}
\usepackage{subfig} 
\usepackage{booktabs}
\usepackage{bm}
\usepackage{algorithm}
\usepackage[noend]{algpseudocode}
\usepackage{pifont}
\newcommand{\cmark}{\ding{51}}%
\newcommand{\xmark}{\ding{55}}%
\usepackage{array}

\usepackage[pagebackref=true,breaklinks=true,colorlinks,bookmarks=false]{hyperref}

\newfloat{algorithm}{t}{lop}

\def\cvprPaperID{6151} 
\def\confYear{CVPR 2021}

\begin{document}

\newcolumntype{M}[1]{>{\centering\arraybackslash}m{#1}}

\newcommand{\vct}[1]{\ensuremath{\boldsymbol{#1}}}
\newcommand{\mat}[1]{\mathtt{#1}}
\newcommand{\set}[1]{\ensuremath{\mathcal{#1}}}
\newcommand{\con}[1]{#1} 
\newcommand{\T}{\ensuremath{^\top}}
\newcommand{\ind}[1]{\ensuremath{\mathbb 1_{#1}}}
\newcommand{\diag}[1]{\mathrm{diag}(#1)}
\newcommand{\cov}[1]{\mathrm{cov}(#1)}
\newcommand{\EX}{\mathbb{E}} 
\newcommand{\targetEmbedding}{r_t}
\newcommand{\domainEmbedding}{e_d}

\newcommand{\inputset}{\set{X}}
\newcommand{\inputSample}{x}
\newcommand{\outputset}{\set{Y}}
\newcommand{\outputSample}{y}
\newcommand{\domainset}{\set{D}}
\newcommand{\domainsetcardinality}{{K}}

\newcommand{\domain}{d}
\newcommand{\targetdomain}{t}
\newcommand{\modelDistribution}{p^{y}_{x}}
\newcommand{\domainDistribution}[1]{p^{y}_{x, {#1}}}
\newcommand{\marginaldomainDistribution}{p^{x}}

\newcommand{\mean}{\mu}
\newcommand{\variance}{\sigma^2}
\newcommand{\batchMean}{\tilde{\mu}_d}
\newcommand{\batchVariance}{{\tilde{\sigma}_d}^2}
\newcommand{\batchStddv}{{\tilde{\sigma}_d}}
\newcommand{\stddv}{\sigma}
\newcommand{\popMean}{\hat{\mu}}
\newcommand{\popVariance}{\hat{\sigma}^2}
\newcommand{\popStddv}{\hat{\sigma}}
\newcommand{\batch}{b}
\newcommand{\targetSample}{{x_t}}

\newcommand{\methodName}{BNE}
\newcommand{\discoveryName}{DNet}
\newcommand{\pacs}{PACS}
\newcommand{\officeT}{Office-31}
\newcommand{\officecaltech}{Office-Caltech}
\newcommand{\alexnet}{AlexNet}
\newcommand{\resnet}{ResNet-18}
\newcommand{\baseline}{DeepAll}

\newcommand{\gain}{$\Delta \%$}
\newcommand{\averageDeepAll}{Avg. DA}
\newcommand{\averagePerfomance}{Avg.}
\newcommand{\averageDomain}{Avg. Domain}
\newcommand{\averageClass}{Avg. Class}

\newcommand{\macroBatch}{T}
\newcommand{\domainBatch}{{\batch_{\domain}}}
\newcommand{\domainBatchCardinality}{{n}}
\newcommand{\layerDomainBatchMean}{\mu_{\domainBatch}^l}
\newcommand{\layerDomainBatchStddv}{\sigma_{\domainBatch}^l}
\newcommand{\domainBatchMean}{\mu_{\domainBatch}}
\newcommand{\domainBatchStddv}{\sigma_{\domainBatch}}
\newcommand{\layerBatchMean}{\batchMean^l}
\newcommand{\layerBatchStddv}{{\batchStddv^l}}
\newcommand{\layerPopMean}{{{\hat{\mu}}_{\domain}^l}}
\newcommand{\layerPopStddv}{{{\hat{\sigma}}_{\domain}^l}}
\newcommand{\domainPred}{f_\domain^\targetdomain}

\title{Batch Normalization Embeddings for Deep Domain Generalization}

\author{Mattia Segu, Alessio Tonioni \& Federico Tombari \\
Google \\
\texttt{\{msegu,alessiot,tombari\}@google.com} \\
}

\maketitle


\input{chapters/0-abstract}

\input{chapters/1-intro}

\input{chapters/2-related}

\input{chapters/3-method_new}

\input{chapters/4-experiments}

\input{chapters/5-conclusions}

{\small
\bibliographystyle{ieee_fullname}
\bibliography{egbib}
}


\clearpage
\input{chapters/6-supplementary.tex}


\end{document}

%% file: chapters/0-abstract.tex
\begin{abstract}
Domain generalization aims at training machine learning models to perform robustly across different and unseen domains.
Several recent methods use multiple datasets to train models to extract domain-invariant features, hoping to generalize to unseen domains. 
Instead, first we explicitly train domain-dependant representations by using ad-hoc batch normalization layers to collect independent domain's statistics.
Then, we propose to use these statistics to map domains in a shared latent space, where membership to a domain can be measured by means of a distance function.
At test time, we project samples from an unknown domain into the same space and infer properties of their domain as a linear combination of the known ones. 
We apply the same mapping strategy at training and test time, learning both a latent representation and a powerful but lightweight ensemble model.
We show a significant increase in classification accuracy over current state-of-the-art techniques on popular domain generalization benchmarks: PACS, Office-31 and Office-Caltech.
%
\end{abstract}

%% file: chapters/1-intro.tex
\section{Introduction}
\input{figures/teaser.tex}
Machine learning models trained on a certain data distribution often fail to generalize to samples from different distributions. 
This phenomenon is commonly referred to in literature as \textit{domain shift between training and testing data}~\cite{sugiyama2007mixture,luo2019taking}, and is one of the biggest limitations of data driven algorithms.
Assuming the availability of few annotated samples from the test domain, the problem can be mitigated by fine-tuning the model with explicit supervision~\cite{yosinski2014transferable} or with domain adaptation techniques~\cite{wang2018deep}.
Unfortunately, this assumption does not always hold in practice as it is often unfeasible in real scenarios to collect samples for any possible environment.
%

%
\textit{Domain generalization} refers to algorithms to solve the \textit{domain shift} problem by learning models robust to unseen domains.
Several works leverage different domains at training time to learn a domain-invariant feature extractor~\cite{muandet2013domain,ghifary2015domain,koch2015siamese,motiian2017unified,li2018domain}.
Other works focus on optimizing the model parameters to obtain consistent performance across domains via ad-hoc training policies~\cite{tobin2017domain,shankar2018generalizing,volpi2018generalizing,li2019episodic}, while a different line of work requires modifications to the model architecture to achieve domain invariance~\cite{khosla2012undoing,li2017deeper,ding2017deep,mancini2018best}.
%

%
%
%
While these methods try to extract domain-invariant features, we go in the opposite direction and explicitly leverage domain-specific representations by collecting domain-dependent batch normalization (BN) statistics for each of the domains available at training time. 
By doing so, we train a lightweight ensemble of domain-specific models sharing all parameters except for BN statistics.
Peculiarly to our proposal, we use the accumulated statistics to map each domain as a point in a \emph{latent space of domains}.
We will refer to this mapping as the Batch Normalization Embedding (\methodName{}) of a domain.
\autoref{fig:teaser} sketches a visualization of such space for the case of three domains available at training time (\eg Photo, Art Painting and Cartoon).  
At convergence, each training domain is mapped to a single point in the domain space. Then, at test time unseen samples from unknown domains can be mapped to the same space by means of their instance normalization statistics.
By measuring the distances between the instance normalization statistics of the test sample (black dot) and the accumulated population statistics of each domain (colored dots), we can infer properties of the unknown test domain.
%
%
%
Specifically, we leverage the reciprocal of such distances at test time to weigh the domain-specific predictions of our lightweight ensemble and accurately classify an unseen sample from an unknown domain.
The same combination of domain-specific models can be used at training time on samples from the known domains to force the ensemble to learn a meaningful latent space and logits that can be linearly combined according to the proposed weighting strategy. 

To sum up the contributions of our work: 
(i) we propose to accumulate domain-specific batch normalization statistics accumulated on convolutional layers to map image samples into a latent space where membership to a domain can be measured according to a distance from domain \methodName{}s;
(ii) we propose to use this concept to learn a lightweight ensemble model that shares all parameters excepts the normalization statistics and can generalize better to unseen domains;
(iii) compared to previous work, we do not discard domain-specific attributes but exploit them to learn a domain latent space and map unknown domains with respect to known ones;
(iv) our method can be applied to any modern Convolutional Neural Network (CNN) that relies on batch normalization layers, and scales gracefully to the number of domains available at training time.

%% file: figures/teaser.tex
\begin{figure}
\centering
\includegraphics[width=\linewidth]{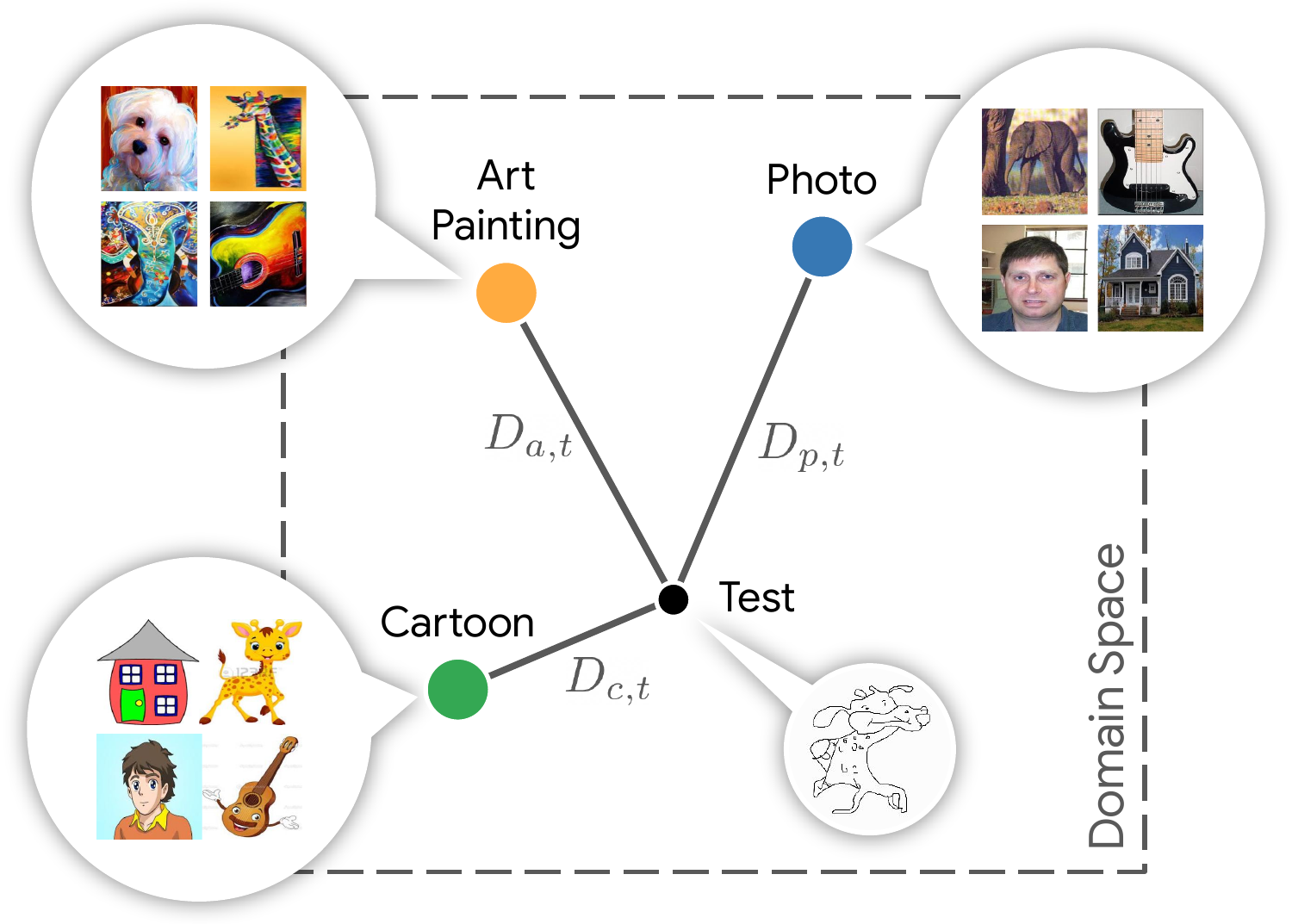}
\caption{Visualization of our method on the PACS dataset when the domains \textit{Art Painting}, \textit{Photo} and \textit{Carton} are available at training time. We propose to use batch normalization layers to implicitly learn a \emph{domain space} onto which map both known (training) and unknown (testing) domains. At test time, we project each target sample independently in the domain space and locate it with respect to the known domains using the corresponding distances $D_{a,t}$, $D_{p,t}$, and $D_{c,t}$. Properties of the unknown domain are revealed by the location of the unseen sample. We leverage these hints to improve classification of each test sample by means of a linear combination of domain specific classifiers, weighted by the inverse of the distances.}
\label{fig:teaser}
\vspace{-0.5cm}
\end{figure}

%

%% file: chapters/2-related.tex
\section{Related Work}

\noindent\textbf{Domain Generalization.}
Most domain generalization works attempt to expose the model to domain shift at training time to generalize to unseen domains. 
Invariance can be encouraged at multiple levels:

\textit{Feature-level,}
denotes methods deriving domain-invariant features by minimizing a discrepancy between multiple training domains.
%
%
Ghifary \etal~\cite{ghifary2015domain} brought domain generalization to the attention of the deep learning community by training multi-task autoencoders to transform images from one source domain into different ones, thereby learning invariant features.
%
%
Analogously, Li \etal~\cite{li2018domain} extended adversarial autoencoders by minimizing the Maximum Mean Discrepancy measure to align the distributions of the source domains to an arbitrary prior distribution via adversarial feature learning.
Conditional Invariant Adversarial Networks~\cite{li2018deep} have been proposed to learn domain-invariant representations, whereas Deep Separation Networks~\cite{bousmalis2016domain} extract image representations partitioned into two sub-spaces: one unique to each domain and one shared. 
Differently, Motiian \etal~\cite{motiian2017unified} propose to learn a discriminative embedding subspace via a Siamese architecture~\cite{koch2015siamese}.
Episodic training~\cite{li2019episodic} was proposed to train a generic model while exposing it to domain shift. 
In each episode, a feature extractor is trained with a badly tuned classifier (or vice-versa) to obtain robust features.
Recently, \cite{matsuura2020domain} proposed a method to simultaneously discover latent domains by clustering featuring together and minimizing feature discrepancy between them.
For all these methods, the limited variety of domains to which the model can be exposed at training time can limit the magnitude of the shift to which the model learns invariance.

\textit{Data-level,}
denotes methods attempting to reduce the training set domain bias by augmenting the cardinality and variety of the samples.
Data augmentation methods based on domain-guided perturbations of input samples~\cite{shankar2018generalizing} or on adversarial examples~\cite{volpi2018generalizing} have been proposed with the purpose of training a model to be robust to distribution shift.
Domain randomization was adopted~\cite{tobin2017domain,loquercio2019deep} to solve the analogous problem of transferring a model from synthetic to real data by extending synthetic data with random renderings.
By performing data augmentation those methods force the feature extractor to learn domain-invariant features, while we argue that discarding domain-specific information might be detrimental for performance.

\textit{Model-based,}
denotes methods relying on ad-hoc architectures to tackle the domain generalization problem. 
\cite{li2017deeper} introduced a low-rank parameterized CNN model, a dynamically parameterized neural network that generalizes the shallow binary undo bias method~\cite{khosla2012undoing}. 
Similarly, a structured low-rank constraint is exploited to align multiple domain-specific networks and a domain-invariant one in~\cite{ding2017deep}.
Mancini \etal~\cite{mancini2018best} train multiple domain-specific classifiers and estimate the probabilities that a target sample belongs to each source domain to fuse the classifiers' predictions.
A recent work~\cite{carlucci2019domain} proposes an alternative approach to tackle domain generalization by teaching a model to simultaneously solve jigsaw puzzles and perform well on a task of interest.
Most of these methods require changes to state-of-the-art architectures, resulting in an increased number of parameters or complexity of the network.

\textit{Meta-learning,}
denotes methods relying on special training policies to train models robust to domain shift.
\cite{li2018learning} extend to domain generalization the widely used model agnostic meta learning framework~\cite{finn2017model}.
\cite{balaji2018metareg} propose a novel regularization function in a meta-learning framework to make the model trained on one domain perform well on another domain.
\cite{huang2020self} propose a training heuristic that iteratively discards the dominant features activated on the training data, challenging the model to learn more robust representations.
A gradient-based meta-train procedure was introduced by \cite{dou2019domain} to expose the optimization to domain shift while regularizing the semantic structure of the feature space.
These methods simulate unseen domains by splitting the training data in a meta-training set and meta-test set, therefore are inherently bounded by the variety of the samples available at training time.

\noindent\textbf{Batch Normalization for distribution alignment.}
The use of separate batch normalization statistics to align a training distribution to a test one has been firstly introduced for domain adaptation~\cite{carlucci2017autodial,carlucci2017just,li2018adaptive}.
The same domain-dependent batchnorm layer has been adapted to the multi-domain scenario~\cite{mancini2018boosting, mancini2018robust} and exploited in a graph-based method~\cite{mancini2019adagraph} that leverages domain meta-data to better align unknown domains to the known ones.
All these works, however, require some representation of the target domain to perform the alignment during training, using either samples or meta-data describing the target domain.
Our approach instead does not rely on any external source of information regarding the target domain.
Domain-specific normalization layers have only recently been proposed for domain generalization in \cite{seo2019learning}, where a cluster of networks is trained to learn an optimal mixture of instance and batch normalization.

%% file: chapters/3-method_new.tex


\section{Method}
The core idea of our method is to exploit domain-specific batch normalization statistics to map known and unknown domains in a shared latent space, where domain membership of samples can be measured according to their distance from the domain embeddings of the known domains.
%

\subsection{Problem Formulation}
\label{ssec:notation}
Let $\inputset$ and $\outputset$ denote the input (\eg images) and the output (\eg object categories) spaces of a model.
Let \mbox{$\domainset = \{\domain_i\}_{i=1}^\domainsetcardinality$} denote the set of the $\domainsetcardinality$ source domains available at training time.
Each domain $\domain_i$ can be described by an unknown conditional probability distribution \mbox{$\domainDistribution{\domain_i}=p(y|x,i)$} over the space $\inputset{} \times \outputset{}$.
The aim of a machine learning model is to learn the probability distribution \mbox{$\modelDistribution=p(y|x)$} of the training set~\cite{bridle1990probabilistic} by training models to learn a mapping $\inputset\rightarrow\outputset$.
We propose to use a lightweight ensemble of models to learn a mapping $(\inputset, \domainset)\rightarrow\outputset$ that leverages the domain label to model a set of conditional distributions $\{\domainDistribution{\domain_i}\}_{i=1}^\domainsetcardinality$, each conditioned on the domain membership.
Let $t$ be a generic target domain available only at testing time and following the unknown probability distribution $\domainDistribution{t}$ over the same space.
Since it is not possible to learn the target distribution $\domainDistribution{t}$ during training, the goal of our method is to accurately estimate it as a mixture (\ie linear combination) of the learned source distributions $\domainDistribution{\domain_i}$.

For each source domain $d \in \domainset$, a training set \mbox{${\set S}_d = \{(x_{1_d}, y_{1_d}), ..., (x_{n_d}, y_{n_d})\}$} containing $n_d$ labelled samples is provided.
The test set \mbox{$\set T = \{x_{1_t}, ..., x_{n_t}\}$} is composed of $m_t$ unlabelled samples collected from the unknown marginal distribution $\marginaldomainDistribution_t$ of the target domain $t$.
As opposed to the domain adaptation setting, we assume that target samples are not available at training time, and that each of them might belong to a different unseen domain.

\subsection{Multi-Source Domain Alignment Layer}
\label{ssec:multi_source_bn}
Neural networks are particularly prone to capture dataset bias in their internal representations~\cite{li2016revisiting}, making internal features distributions highly domain-dependent.
To capture and alleviate the distribution shift that is inherent in the multi-source setting, we draw inspiration from~\cite{carlucci2017just,mancini2018boosting, mancini2018robust, seo2019learning} and adopt batch normalization layers~\cite{ioffe2015batch} to normalize the activations of each domain to the same reference distribution via \emph{domain-specific normalization statistics}.
%
%

At inference time, the activations of a certain domain $\domain$ are normalized by matching their first and second order moments, nominally $(\mean_\domain{}, \variance_\domain{}$), to those of a reference Gaussian with zero mean and unitary variance:
\begin{align} \label{eq:batch_normalization}
    BN(z;\domain{}) = \frac{z - \mean_\domain{}}{\sqrt{\variance_\domain{} + \epsilon}},
\end{align}
where $z$ is an input activation extracted from the marginal distribution $q^z_d$ of the activations from the domain $d$; \mbox{$\mu_d = \EX_{z \sim q^z_d}[z]$} and \mbox{$\variance_d = Var_{z \sim q^z_d}[z]$} are the population statistics for the domain $d$, and $\epsilon>0$ is a small constant to avoid numerical instability.
At training time, the layer collects and applies domain-specific batch statistics $(\batchMean, \batchVariance)$, while updating the corresponding moving averages to approximate the domain population statistics.
%

At inference time, if the domain label $\domain$ of a test sample is unknown or it does not belong to \domainset{}, we can still rely on normalization by \emph{instance statistics}, \ie the degenerate case of batch statistics with batch size equal to $1$.
%
\input{figures/method.tex}
%
\autoref{fig:method} (a) depicts the functioning of a multi-source domain generalization layer.
Our method builds on the observation that for convolutional layers instance statistics and batch statistics are approximations of the same underlying distribution with different degrees of noise.
Since the population statistics are a temporal integration of the batch statistics, the validity of this statement extends to the comparison with them.
For example, statistics for a single channel in the case of a batch normalization layer applied on a 2D feature map of size $H \times W$, for a generic batch size $B$ (\textit{batch statistics}) and for $B=1$ (\textit{instance statistics}), are computed as:
\begin{equation} \label{eq:batch_statistics}
\begin{split}
    \tilde{\mean} &= \frac{1}{B \cdot H \cdot W} \sum_{b,h,w} z_{b,h,w} \\
    &\stackrel{(B=1)}{=} \frac{1}{H \cdot W} \sum_{h,w} z_{h,w} 
\end{split}
\end{equation} 

\begin{equation}
\begin{split}
    \tilde{\stddv}^2 &= \frac{1}{B \cdot H \cdot W} \sum_{b,h,w} (z_{b,h,w} - \tilde{\mean})^2 \\
    &\stackrel{(B=1)}{=} \frac{1}{H \cdot W} \sum_{h,w} (z_{h,w} - \tilde{\mean})^2,
\end{split}    
\end{equation}
where $\tilde{\mean}$ and ${\tilde{\stddv}}^2$ are respectively the batch mean and variance and $z_{b,h,w}$ is the value of a single element of the feature map.
If we consider $z_{b,h,w}$ to be described by a normally distributed random variable $Z \sim \set N(\mean,{\stddv}^2)$, then the instance and batch statistics are an estimate of the parameters of the same gaussian computed over a different number of samples, $H \cdot W$ and $B \cdot H \cdot W$ respectively.
%
In the next section, we explain how we exploit this property to map source domains and unseen samples from unknown domain into the same latent space.

\subsection{Domain Localization in the Batchnorm Latent Space} 
\label{ssec:projection}
The domain alignment layers described in~\autoref{ssec:multi_source_bn} allow to learn the multiple source distributions $\{\domainDistribution{d}\}_{d \in \set D}$ distinctly.
By leveraging them, we can learn a lightweight ensemble of domain-specific models, where every network shares all the weights except for the normalization statistics.
Since such a lightweight ensemble nicely embodies the multiple source distributions, we propose to reduce the domain shift on the unknown target domain by interpolating across these distributions to estimate the unknown distribution $\domainDistribution{t}$.
The resulting target distribution is a weighted mixture of the distributions in the ensemble, for which the choice of the weights depends on the \emph{similarity} of a test sample to each source domain.

%
We denote with a $l \in \set B = \{1, 2, ..., L\}$ in superscript notation the different batch normalization layers in the model.
For each of them we can define a latent space $\set L^l$ spanned by the activation statistics at the $l-th$ layer of the model.
In this space, we observe that single samples $x$ are mapped via their instance statics $(\tilde{\mean},\tilde{\stddv}^2)$, whereas the population statistics accumulated for each domain $(\mean_\domain{}, \variance_\domain{})$ are used to represent domain centroids.
\autoref{fig:method}-(b) shows a visualization of the latent space $\set L^1$ for the PACS dataset~\cite{li2017deeper}, composed of $4$ domains, and $3$ of which (\eg \textit{Art Painting}, \textit{Cartoon}, \textit{Photo}) assumed available at training time. 
Population (big green dots) and instance (small dots) statistics are respectively used to project domains and individual samples. 
We rely on t-SNE\cite{maaten2008visualizing} to visualize instance and population statistics of source and target samples, and we observe how the latent space that we propose allows a spontaneous and stark division between domain clusters.
%
Considering all latent spaces at different layers, we define a batch normalization embedding (\methodName{}) for a certain domain $\domain$ as the stacking of the population statistics computed at every layer:
\begin{align} \label{eq:domain_embedding}
    \domainEmbedding &= [\domainEmbedding^1, \domainEmbedding^2, ..., \domainEmbedding^L] \\ 
                     &= [(\mean_d^1,{\stddv_d^1}^2), (\mean_d^2,{\stddv_d^2}^2), ..., (\mean_d^L,{\stddv_d^L}^2)]. \notag
\end{align}
%
For a target sample $\targetSample$ from an unknown domain $t$, we can derive a projection to the same space by forward propagating it through the network and computing its instance statistics.
The latent embedding $\targetEmbedding$ of $\targetSample$ is defined as the stacked vector of its instance statistics at different batch normalization layers in the network:
\begin{align} \label{eq:target_embedding}
    \targetEmbedding &= [\targetEmbedding^1, \targetEmbedding^2, ..., \targetEmbedding^L] \\
    &= [(\mean_t^1,{\stddv_t^1}^2), (\mean_t^2,{\stddv_t^2}^2), ..., (\mean_t^L,{\stddv_t^L}^2)]. \notag
\end{align}
Each $\targetEmbedding^l$ represents the instance statistics collected at a certain layer $l$ during forward propagation and can be used to map the sample $\targetSample$ in the latent space $\set L^l$ of layer $l$.
%
%
Once the \methodName{} for the test sample is available, it is possible to measure the \textit{similarity} of a target sample $\targetSample$ to one of the known domains $\domain$ as the inverse of the distance between $\targetEmbedding$ and $\domainEmbedding$. 
By extension, this allows a soft 1-Nearest Neighbour domain classification of any test sample.

To compute a distance between two points in $\set L^l$, we consider the means and variances of the corresponding batch normalization layer as the parameters of a multivariate Gaussian distribution. 
We can hence adopt a distance on the space of probability measures, \ie a symmetric and positive definite function that satisfies the triangle inequality.
We select the \textit{Wasserstein distance} for the special case of two multivariate gaussian distributions, but we report a comparison to alternative distances in the supplementary material.
Let $p \sim \set N (\mu_p , C_p)$ and $q \sim \set N (\mu_q , C_q)$ be two normal distributions on $R^n$, with expected value $\mu_p$ and $\mu_q \in R^n$ respectively and $C_p$, $C_q \in R^{n \times n}$ covariance matrices.
Denoting with $||\cdot||_2$ the Euclidean norm on $R^n$, the \textit{2-Wasserstein distance} is:
\begin{align} \label{eq:wasserstein_distance}
  \set W (p, q) &= \phi ((\mu_p , C_p), (\mu_q , C_q)) \\
  &= ||\mu_p - \mu_q||_2^2 + Tr(C_p + C_q - 2(C_q^{\frac{1}{2}}C_pC_q^{\frac{1}{2}})^{\frac{1}{2}}), \notag
\end{align}
$Tr$ being the trace of the matrix.
We rely on \autoref{eq:wasserstein_distance} to measure the distance between a test sample $\targetSample$ and the domain $d$ by summing over the batch normalization layers $l \in \set B$ the distance between the activation embeddings $\targetEmbedding^l$ and $\domainEmbedding^l$:
%
%
%
%
\begin{align} \label{eq:total_distance}
  D_{\set L}(\domainEmbedding, \targetEmbedding) &= \sum_{l \in \set B} \set W (\domainEmbedding^l, \targetEmbedding^l) \\
  &= \sum_{l \in \set B} \phi ((\mean_d^l , Diag({\stddv_d^l}^2)), (\mean_{\targetSample}^l , Diag({\stddv_{\targetSample}^l}^2))). \notag
\end{align}
%
\autoref{eq:batch_statistics} shows that instance and batch statistics differ only for the number of samples over which they are estimated, making the comparison meaningful.
The \textit{similarity} of a test sample $\targetSample$ to the domain $d$ is defined as the reciprocal of the distance from that domain and denoted as $w^\targetdomain_d$.
%
%

Once the similarity to each source domains is computed, we can use them to recover the unknown target distribution $\domainDistribution{t}$ as a mixture (\ie a linear combination) of the learned source distributions $\domainDistribution{d}$ weighted by the corresponding domain similarity:
\begin{align} \label{eq:target_mixture}
    \domainDistribution{t} = \frac{\sum_{d \in \set D} w^\targetdomain_d \domainDistribution{d}}{\sum_{d \in \set D} w^\targetdomain_d}.
\end{align}
We denote with $f(\cdot)$ the result of a forward pass in a neural network.
We get the final prediction of our lightweight ensemble model $f(\targetSample)$ as a linear combination of the domain dependant models $f(\targetSample|d)$:
\begin{align} \label{eq:weighted_prediction}
  f(\targetSample) = \frac{\sum_{d \in \set D} w^\targetdomain_d f(\targetSample | \domain{})}{\sum_{d \in \set D} w^\targetdomain_d},
\end{align}
%
%
%
%
\autoref{fig:method}-(b) shows a visualization of our localization technique superimposed over the t-SNE plot of instance and population statistics.
By measuring the distance of a test samples (yellow dot) from the training domain embeddings (big green dots), we obtain ad-hoc mixture coefficients for each test sample. 
%

Our formulation allows to navigate in the latent space of the batchnorm statistics.
Specifically, if a test sample belongs to one of the source domains, our method assigns a high weight to the prediction of the corresponding domain-specific model.
On the other hand, if the test sample does not belong to any of the source domains, the final prediction will be expressed as a linear combination of the domain-dependant models embodied in our lightweight ensemble.

\subsection{Training Policy}
\label{ssec:training}
%
To encourage a well-defined latent space for every batch normalization layer, we replicate at training time the distance weighting procedure described in~\autoref{eq:weighted_prediction} to compute predictions on samples from known domains.
Each training batch is composed of $\domainsetcardinality$ domain batches with an equal number of samples.
During every training step, (i) the domain batches are first propagated to update the corresponding domain population statistics $(\mean_\domain, \variance_\domain,)$.
Then, (ii) all individual samples are propagated assuming an unknown domain to collect their instance statistics and compute the domain similarities $w^\targetdomain_d$, as in~\autoref{ssec:projection}.
Finally, (iii) each sample is propagated under $\domainsetcardinality$ different domain assumptions (\ie through the corresponding domain-specific branches) and the resulting domain-specific predictions are weighted according to~\autoref{eq:weighted_prediction}.
Applying this procedure during training promotes the creation of a well-defined latent space.

Since we initialize our model with weights pre-trained on ImageNet~\cite{deng2009imagenet}, each domain-specific batch normalization branch needs to be specialized before starting the distance training (DT) procedure described above, otherwise convergence problems might occur.
We thus \textit{warm-up} domain-specific batch normalization statistics by pre-training the model on the whole dataset following the standard procedure, except for the accumulation and application of domain-specific batch normalization statistics.

%% file: figures/method.tex
%
\begin{figure*} [t]
\centering
\subfloat[Multi-Source Domain Alignment Layer.]{\raisebox{0.2cm}{{\includegraphics[trim=0cm 0cm 14.0cm 5.4cm, clip, width=0.41\linewidth]{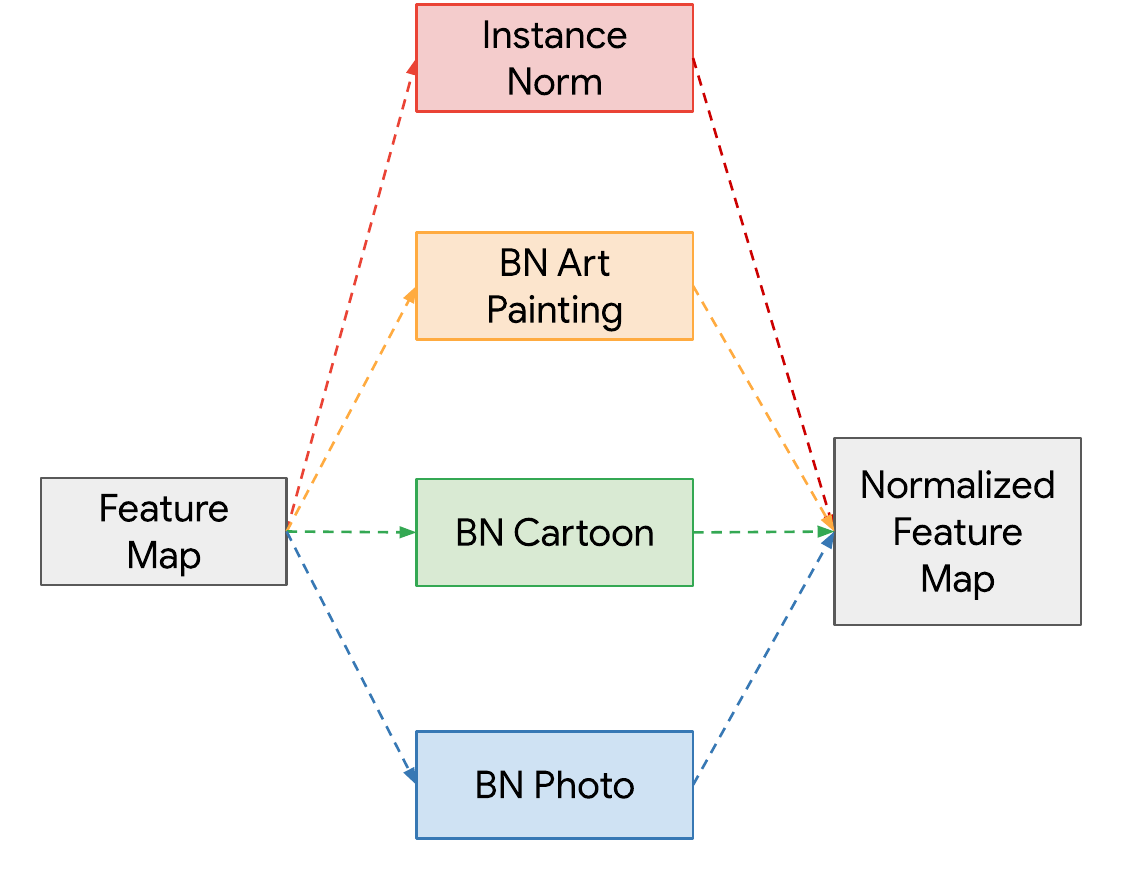}}}}
\quad
\subfloat[t-SNE plot of instance and population statistics for seen and unseen domains.]{{\includegraphics[width=0.53\linewidth]{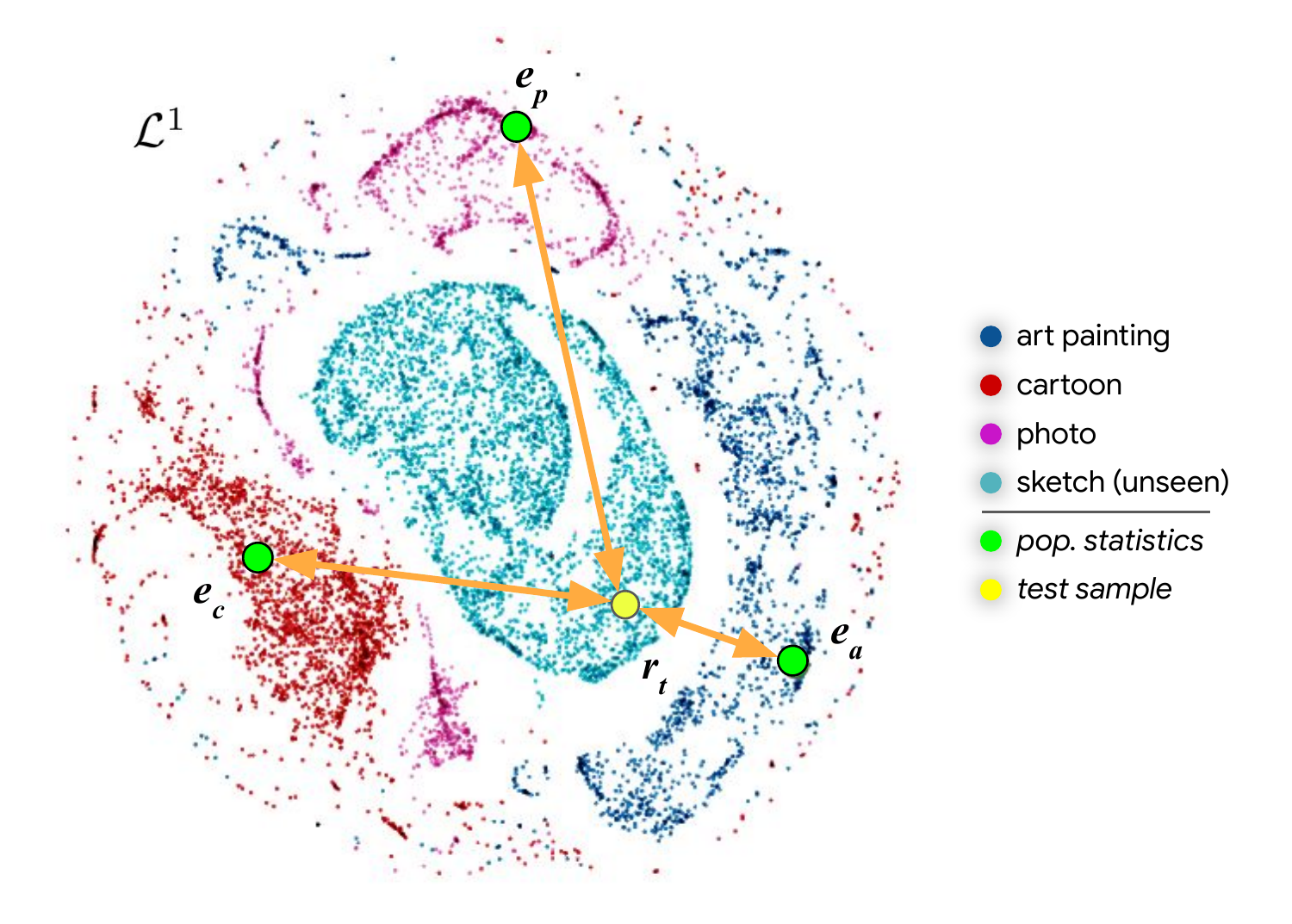}}} 

\caption{Our method on PACS~(\cite{li2017deeper}) with \textit{Sketch} as unknown domain. A Multi-Source Domain Alignment Layer (a) collects domain-specific population statistics and compute instance statistics for test samples. After training, the population and instance statistics map respectively the source domains and the test samples into a latent space, where domain similarity can be measured by distances between embedding vectors. In (b), we visualize the learned domain space $\set L^1$ by means of a t-SNE plot of instance normalization and population statistics for a model trained with our method. Each test sample from the unseen domain \textit{sketch} can be localized through its instance statistics (cyan dots) with respect to the known domains, embedded by the population statistics (green dots). Considering a test sample embedding, \eg $r_t$, the estimated distances (orange arrows) will be used to weigh the predictions of domain-specific classifiers.}
\label{fig:method}
\vspace{-0.3cm}
\end{figure*}
%

%% file: chapters/4-experiments.tex

\section{Experiments}
\subsection{Experimental Settings}
By means of a synecdoche, we name our method after \methodName{}, its main component.

\noindent\textbf{Datasets.}
We evaluate our method on three domain generalization benchmarks: 

\noindent\textit{\pacs{}}~\cite{li2017deeper} features $4$ domains (\textit{Art Painting}, \textit{Cartoon}, \textit{Photo}, \textit{Sketch}) with a significant domain shift.
Each domain includes samples from 7 different categories, for a total of $9991$ samples.
Some examples are shown in \autoref{fig:method}.

\noindent\textit{\officeT{}}~\cite{saenko2010adapting} was originally introduced for domain adaptation and has been subsequently used for domain generalization. 
The dataset is composed of $3$ different sources and $31$ categories, representing images captured with a Webcam and a dSLR camera or collected from the Amazon website.

\noindent\textit{\officecaltech{}}~\cite{gong2012geodesic} is a variant of \officeT{} featuring one additional domain, derived from the Caltech-256 dataset~\cite{griffin2007caltech}.
The dataset is composed of the $10$ categories shared between Caltech-256 and the domains in \officeT{}.

\noindent\textbf{Evaluation Protocol.}
Coherently with other works, we evaluate both the \alexnet{}~\cite{krizhevsky2012imagenet} and the  \resnet{}~\cite{he2016deep} architectures.
For the experiments on \pacs{} and \officeT{} we follow the standard \textit{leave-one-domain-out} evaluation procedure, where the model is trained on all domains but one, and tested on the left-out one.
For \officecaltech{} we do the same but also test following a \textit{leave-two-domain-out} procedure.
Since the original version of \alexnet{} does not include batch normalization layers, we adopt a variant with batch normalization applied on the activations of each convolutional layer~\cite{simon2016imagenet}.
Since the goal of domain generalization is to leverage multiple sources to learn models that are robust on any target domain, the natural deep-learning baseline to compare against consists in training directly on the merged set of source domains.
We will refer to it as (\baseline{}).
We compare our method against this strong baseline and several deep-learning based state-of-the-art methods for domain generalization.
Since different methods rely on different initializations of network weights, which result in different baselines, we compare with methods providing their own baseline and report for every competitor: the performance on each unseen domain, the average baseline performance (\averageDeepAll{}), the average performance of the method itself (\averagePerfomance{}) and the relative gain (\gain{}). 

\subsection{Domain Generalization for Classification}
We compare \methodName{} against several methods for object classification on commonly used benchmarks.
Additional experiments are reported in the supplementary material.
%
%
\input{tables/SOTA-PACS-Resnet.tex}
\noindent{}\textbf{PACS.} We first benchmark our method on the \pacs{} dataset~\cite{li2017deeper}, which presents a challenging domain generalization setting for object recognition.
Every test uses $3$ domains as training set and one as unknown test set; for each of this leave-one-out configurations, we train a model from the same initialization for $60$ epochs.
We test our method using the \resnet{} architecture and report the results in \autoref{table:SOTA-PACS-Resnet}.
Overall, our proposal obtains the best absolute accuracy on 1 out of 4 target sets, with the second best average accuracy (\averagePerfomance{}) of $83.1\%$ and a relative gain (\gain{}) of $+5.86$ making it the second most effective algorithm on this dataset.
Since all the networks are initialized with weights trained on ImageNet, they are implicitly biased towards the \textit{Photo} domain, as testified by the higher accuracy on it when treated as test set.
\textit{Sketch}, instead, is arguably the more challenging domain, as testified by the lower accuracy achieved by all methods. 
It is in this scenario that our method is able to provide the bigger gain ($+9.6\%$ absolute gain in accuracy over our baseline).
The best perfomance on this datasets is obtained by DSON~\cite{seo2019learning}, which proposes to learn an ad-hoc mixture of instance and batch normalization statics to improve generalization.
This characteristic makes it a perfect candidate to be extended with the domain embedding strategy that we propose in \autoref{ssec:projection}. An in-depth discussion on this topic is provided in \autoref{ssec:distance}.


\input{tables/SOTA-Office-31-Alexnet.tex}
%
\noindent{}\textbf{Office-31.} We evaluate our method on \officeT{}~\cite{saenko2010adapting} and follow the \textit{leave-one-domain-out} protocol.
To compare with published results, we use \alexnet{} initialized with ImageNet weights and train it with our method for $100$ epochs with learning rate $10^{-4}$.
\autoref{table:SOTA-Office-31-Alexnet} shows that our approach obtains the best absolute accuracy in two out of three test scenarios and a relative gain comparable or better than the alternatives.
The \textit{Amazon} target domain proves to be the most challenging setting, as the images are acquired in ideal conditions (\ie white backgrounds, studio lighting\dots) that are fairly different from the ImageNet domain.
In this challening generalization scenario, \methodName{} boosts the absolute accuracy with respect to the baseline by an impressive $+11.2\%$ and $+2.3\%$ with respect to the closest competitor.   
%

\input{tables/SOTA-Office-Caltech-Alexnet.tex}
\noindent{}\textbf{Office-caltech.} We evaluate our method also on \officecaltech{}~\cite{gong2012geodesic} and follow the standard evaluation procedure for this dataset, enumerating cases with a single target domain, either \textit{Amazon} or \textit{Caltech}, and scenarios with pairs of target domains: \textit{Dslr-Webcam} and \textit{Amazon-Caltech}. 
We use \alexnet{} initialized with ImageNet weights to compare with published results, and train \methodName{} for $100$ epochs.
\autoref{table:SOTA-Office-Caltech-Alexnet} shows that our approach achieves the best average accuracy and the best gain with respect to the baseline.
The performance boost delivered with our method is especially evident in the challenging scenario where $2$ domains are treated as targets, \eg $+6.4\%$ absolute accuracy on \textit{Dslr-Webcam}. 

\subsection{Ablation Study}
\label{ss:ablation}
\input{tables/method.tex}
%
To measure the impact on performance of the different components of our method, we run ablation experiments on the \pacs{} dataset using the \resnet{} backbone, and report the results in \autoref{table:method}.
We compare again with the \baseline{} baseline and against a \textit{DiscoveryNet} (DNet, row (d)), a variant of \methodName{} inspired by \cite{mancini2018robust}. 
While in \methodName{} we propose to assign domain membership by looking at the distance between the latent batch normalization embeddings, with DNet domain membership is learned through a domain classification network.
On row (a) we show the performance gain attributable to the usage of separate batchnorm statistics for each training domain, while using at inference time the projection and weighting strategy described in \autoref{ssec:projection}; row (b) extends the this approach by leveraging \textit{distance weighting} (DT) \textit{at training time}, as described in \autoref{ssec:training}; finally, row (c) includes a warm-up phase in the initial training phase to help population statistics to converge to stable values before starting distance training.
By comparing the average accuracy (\averagePerfomance{}) across the four possible target sets, it is clear how every component contributes to an increase in performance with respect to the baseline.
By comparing line (c) to (d), we can notice how our proposal is more effective than the variant DNet inspired by the domain mapping strategy from~\cite{mancini2018robust}, while also requiring less parameters.
More details on DNet and extensive comparisons are reported in the supplementary material.


\subsection{Choosing a Distance Metric} \label{ssec:distance}
\input{tables/distance.tex}
%
A crucial component of our method is the distance function, used to locate each test sample with respect to the known domains.
As mentioned in \autoref{ssec:projection}, we picked the Wasserstein distance for this task after a detailed preliminary study among different options, reported in \autoref{table:distance}.
We considered three different distances: using a fixed value for the distance (\textit{Uniform}), equivalent to averaging predictions of domain specific branches; using the Bhattacharya distance; using the Wasserstein distance.
The basic Uniform distance setting is similar to \cite{seo2019learning}, with the main difference being the normalization layer used: a domain-specific batchnorm in our case, a learned mixture of instance and batch normalizations for \cite{seo2019learning}.
Their approach thus also require to learn additional parameters.
In our experiments, the Wasserstein distance proves to be a more principled choice, consistently delivering the best performance both on average and on any specific left-out domain.
Measuring the similarity of all samples to one of the training domain by means of a distance function is always more effective than blindly averaging predictions, therefore the effectiveness of our method derives from the accurate sample-wise domain attribution rather than from the chosen normalization layer.
This observation leaves open the opportunity of combining our distance weighting scheme with other recently proposed normalization methods, \eg~\cite{seo2019learning}.

%% file: tables/SOTA-PACS-Resnet.tex
\setlength{\tabcolsep}{4pt}
\begin{table*}[t]
\begin{center}
\caption{State-of-the-art comparison on \pacs{} with \resnet{}. Methods with * do not use domain labels.}
\label{table:SOTA-PACS-Resnet}
\begin{tabular}{c|c c c c|c c c}
\hline\noalign{\smallskip}
Method & Art & Cartoon & Photo & Sketch & \averageDeepAll{} & \averagePerfomance{} & $\Delta \%$ \\
\noalign{\smallskip}
\hline
\noalign{\smallskip}
CrossGrad -~\cite{shankar2018generalizing} & 78.7 & 73.3 & 94.0 & 65.1 & 79.1 & 77.8 & -1.64 \\ 
MetaReg -~\cite{balaji2018metareg} &  79.9 & 75.1 & 95.2 & 69.5 & 79.9 & 81.7 & +2.25 \\ 
MLDG -~\cite{li2018learning} &  79.5 & 77.3 & 94.3 & 71.5 & 79.1 & 80.7 & +2.02 \\ 
Epi-FCR -~\cite{li2019episodic} &  82.1 & 77.0 & 93.9 & 73.0 & 79.1 & 81.5 & +3.03 \\ 
JiGen* -~\cite{carlucci2019domain} &  79.4 & 75.3 & \textbf{96.0} & 71.4 & 79.1 & 80.5 & +1.77 \\ 
MASF -~\cite{dou2019domain} &  80.3 & 77.1 & 94.99 & 71.69 & 79.2 & 81.0 & +1.01 \\ 
D-SAM - \cite{d2018domain} & 77.3 & 72.4 & 95.3 & 77.8 & 79.5 & 80.7 & +1.47 \\
MMLD* - \cite{matsuura2020domain} & 81.3 & 77.2 & 96.1 & 72.3 & 78.7 & 81.8 & +3.93 \\
DSON - \cite{seo2019learning} & \textbf{84.7} & 77.6 & 95.9 & \textbf{82.2} & 78.9 & \textbf{85.1} & \textbf{+7.85} \\
\noalign{\smallskip}
\hline
\noalign{\smallskip}
DeepAll &  75.8 & 73 & 94.4 & 70.9 & - & 78.5 & - \\
\methodName{} (\textit{Ours}) &  78.8 & \textbf{78.9} & 94.8 & 79.7 & 78.5 & 83.1 & +5.86 \\ 
\hline
\end{tabular}
\end{center}
\end{table*}
%

%% file: tables/SOTA-Office-31-Alexnet.tex
%
\setlength{\tabcolsep}{4pt}

\begin{table*}[t]
\begin{center}
\caption{State-of-the-art comparison on Office-31 with~\alexnet{}.}
\label{table:SOTA-Office-31-Alexnet}
\begin{tabular}{c|c c c|c c c}
\hline\noalign{\smallskip}
Method & Amazon & Dslr & Webcam & \averageDeepAll{} & \averagePerfomance{} & $\Delta \%$ \\
\noalign{\smallskip}
\hline
\noalign{\smallskip}
UB -~\cite{khosla2012undoing} & 42.4 & 98.5 & 93.4 & 74.2 & 78.1 & +5.26 \\
DSN -~\cite{bousmalis2016domain}  & 44.0 & 99.0 & 94.5 & 74.2 & 79.2 & +6.74 \\
MTAE -~\cite{ghifary2015domain}  &  43.7 & 99.0 & 94.2 & 74.2 & 79.0 & +6.47 \\
DGLRC -~\cite{ding2017deep}  &  45.4 & \textbf{99.4} & \textbf{95.3} & 74.2 & 80.0 & +7.82 \\
MCIT -~\cite{rahman2019multi}  &  51.7 & 97.9 & 94.0 & 74.2 & 81.2 & \textbf{+9.43} \\
\noalign{\smallskip}
\hline
\noalign{\smallskip}
DeepAll & 43.8 & 94.1 & 88.4 & - & 75.4 & - \\
\methodName{} (\textit{Ours}) &  \textbf{54.0} & \textbf{99.4} & 92.3 & 75.4 & \textbf{81.9} & +8.62 \\
\hline
\end{tabular}
\end{center}
\vspace{-0.3cm}
\end{table*}

%

%% file: tables/SOTA-Office-Caltech-Alexnet.tex
%
\setlength{\tabcolsep}{4pt}
\begin{table*}[t]
\begin{center}
\caption{State-of-the-art comparison on Office-Caltech with~\alexnet{}.}
\label{table:SOTA-Office-Caltech-Alexnet}
\scalebox{0.95}{
\begin{tabular}{c|c c c c|c c c}
\hline\noalign{\smallskip}
Method & Amazon & Caltech & Dslr,   & Amazon, & \averageDeepAll{} & \averagePerfomance{} & $\Delta \%$ \\
       &        &         &  Webcam & Caltech &         &         &   \\
\noalign{\smallskip}
\hline
\noalign{\smallskip}
UB -~\cite{khosla2012undoing} &  91.0 & 86.0 & 80.5 & 70.0 & 84.5 & 81.9 & -3.08 \\
DSN -~\cite{bousmalis2016domain}  &  - & - & 85.8 & 81.2 & 84.5 & - & - \\
MTAE -~\cite{ghifary2015domain}  &  93.1 & 86.2 & 85.3 & 80.5 & 84.5 & 86.3 & +2.13 \\
DGLRC -~\cite{ding2017deep}  &  \textbf{94.2} & \textbf{87.6} & 86.3 & 82.2 & 84.5 & 87.6 & +3.67 \\
MDA -~\cite{hu2019domain}  &  93.5 & 86.9 & 84.9 & 82.6 & 84.5 & 87.0 & +2.96 \\
CIDG -~\cite{li2018conditional}  &  93.2 & 85.1 & 83.7 & 65.91 & 84.5 & 82.0 & -2.96 \\
MCIT -~\cite{rahman2019multi}   &  93.3 & 86.3 & 85.2 & 82.7 & 84.5 & 86.9 & +2.84 \\
\noalign{\smallskip}
\hline
\noalign{\smallskip}
DeepAll & 91.7 & 82.1 & 83.4 & 84.5 & - & 85.4 & - \\
\methodName{} (\textit{Ours}) &  93.7 & 85.9 & \textbf{89.8} & \textbf{87.7} & 85.4 & \textbf{89.3} & \textbf{+4.57} \\
\hline
\end{tabular}
}
\end{center}
\end{table*}
\setlength{\tabcolsep}{1.4pt}
%

%% file: tables/method.tex
\setlength{\tabcolsep}{1.5pt}
\begin{table*}[t]
\begin{center}
\caption{Comparison of different variants of our method on PACS with Resnet.}
\label{table:method}
\begin{tabular}{M{1.5cm}|M{0.8cm}M{1.4cm}|M{1.2cm}M{1.2cm}M{1.2cm}M{1.2cm}|M{1.2cm}M{1.2cm}}
\hline\noalign{\smallskip}
Method & DT & Warm-up & Art & Cartoon & Photo & Sketch & Avg. & $\Delta \%$ \\
\noalign{\smallskip}
\hline
\noalign{\smallskip}
DeepAll & - & - & 75.8 & 73.0 & 94.4 & 70.9 & 78.5 & - \\
\midrule
(a) \methodName{} & \xmark & \xmark & 74.7 & 71.7 & 93.0 & 74.8 & 78.6 & +0.03 \\
(b) \methodName{} & \cmark & \xmark & \textbf{79.8} & 76.0 & 92.5 & 72.5 & 80.2 & +2.13 \\
(c) \methodName{} & \cmark & \cmark & 78.8 & \textbf{78.9} & \textbf{94.8} & \textbf{79.7} & \textbf{83.1} & \textbf{+5.86} \\
\noalign{\smallskip}
\hline
\noalign{\smallskip}
(d) \discoveryName{} & \cmark & \cmark & 77.3 & 73.8 & 94.2 & 71.2 & 79.1 & +0.08 \\
\hline
\end{tabular}
\end{center}
\end{table*}
\setlength{\tabcolsep}{1.4pt}
%

%% file: tables/distance.tex
\setlength{\tabcolsep}{4pt}
\begin{table*} [h!]
\begin{center}
\caption{Comparison of different distance measures with our method on PACS with Alexnet.}
\label{table:distance}
\begin{tabular}{M{1.8cm}|M{2.1cm}|M{1.2cm}M{1.2cm}M{1.2cm}M{1.2cm}|M{1.2cm}}
\hline\noalign{\smallskip}
Method & Distance & Art & Cartoon & Photo & Sketch & Average \\
\noalign{\smallskip}
\hline
\noalign{\smallskip}
DeepAll & - & 64.4 & 65.4 & 88.0 & 53.8 & 67.9 \\
\noalign{\smallskip}
\hline
\noalign{\smallskip}
\methodName{} (\textit{Ours}) & Uniform & 64.9 & 64.0 & 88.7 & 61.7 & 69.8 \\
\methodName{} (\textit{Ours}) & Bhattacharyya & 66.3 & 64.6 & 89.4 & 64.3 & 71.2 \\
\methodName{} (\textit{Ours}) & Wasserstein & \textbf{66.7} & \textbf{65.7} & \textbf{89.5} & \textbf{66.8} & \textbf{72.2} \\
\hline
\end{tabular}
\end{center}
\vspace{-0.3cm}
\end{table*}
\setlength{\tabcolsep}{1.4pt}
%

%% file: chapters/5-conclusions.tex
\section{Conclusions}
Our method allows to navigate in the latent space of batch normalization statistics, describing unknown domains as a combination of the known ones.
We rely on domain-specific normalization layers to disentangle independent representations for each training domain, and then use such implicit embeddings to localize unseen samples from unknown domains.
%
%
%
Our method outperforms many alternatives on a number of domain generalization benchmarks~(\cite{li2017deeper,saenko2010adapting,gong2012geodesic}), underlining the advantage of maintaining specific domain representations over forcing invariant representations.
%
We believe that our work highlights interesting properties of batch normalization layers, not extensively explored yet. 
Our formulation could also be easily extended to the domain adaptation setting by injecting unlabelled samples from the target domain during training.
If few target samples happened to be available at the same time, they could all be used to retrieve a less biased estimate of the statistics of the unseen domain.
We plan to explore these directions in future work.

%% file: chapters/6-supplementary.tex
\newcommand{\refColor}{blue}

\section{Supplementary Material}
We provide supplementary material to further validate our method and complement the experimental section included in the main paper.
\autoref{ssec:algo} provides an algorithmic overview of the proposed training policy and additional training details are reported in \autoref{ssec:training_settings};
\autoref{ssec:additional_results} shows additional results obtained with \resnet{}~\cite{he2016deep} on \officeT{}~\cite{saenko2010adapting} and \officecaltech{}~\cite{gong2012geodesic}, and with \alexnet{} on \pacs{}~\cite{li2017deeper};
In \autoref{ssec:ablation}, we conduct a qualitative analysis to verify our choices in terms of batch sizes and distance measures. Moreover, we validate \methodName{} against other popular normalization strategies.
In \autoref{ssec:latentSpaceExperimental}, we validate quantitatively our latent space proposal.
Finally, we extensively compare the performance of \methodName{} against the variant \discoveryName{} that leverages a domain discovery network.

\textit{N.B.:} Blue references point to the original manuscript.

\subsection{Training Policy} \label{ssec:algo}
We here provide a formalization of the distance training policy described in~\textcolor{\refColor}{Sec. 3.4}.


Let $\macroBatch = \{\domainBatch\}_{\domain \in \domainset}$ be a training batch composed of $\domainsetcardinality$ domain batches, each containing $\domainBatchCardinality$ samples from the corresponding domain $\domain$: $\domainBatch = \{(x_d^i,y_d^i)\}_{i=1}^\domainBatchCardinality$.
In \autoref{algobox}, we illustrate the training procedure for a single training batch $\macroBatch$ using the same notation as in the original manuscript.

\begin{algorithm*}[h] 
\caption{Training Step for a batch \macroBatch} 
\label{algobox}
\begin{algorithmic}[1]
  \For{$\domainBatch \in \macroBatch$}\Comment{\textbf{for every domain batch}}
    \State Collect domain batch statistics $(\batchMean, \batchStddv)$\Comment{forward propagation}
    \State $\layerPopMean \longleftarrow0.99\layerPopMean + 0.01\layerBatchMean \quad \forall l \in \set B$\Comment{update domain population mean}
    \State $(\layerPopStddv)^2 \longleftarrow 0.99(\layerPopStddv)^2 + 0.01(\layerBatchStddv)^2 \quad \forall l \in \set B$\Comment{update domain population variance}
    \State $\domainEmbedding^l \longleftarrow (\layerPopMean,(\layerPopStddv)^2) \quad \forall l \in \set B$\Comment{update domain layer embeddings}
    \State $\domainEmbedding \longleftarrow [\domainEmbedding^1, \domainEmbedding^2, ..., \domainEmbedding^L]$\Comment{update domain embedding}
      
  \EndFor
  \For{$(\targetSample,y_t) \in \macroBatch$}\Comment{\textbf{for every sample}}
      \State Collect instance statistics $(\mean_t,{\stddv_t}^2)$\Comment{forward propagation}
      \State $\targetEmbedding^l \longleftarrow (\mean_t^l,{\stddv_t^l}^2) \quad \forall l \in \set B$\Comment{define target layer embeddings}
      \State $\targetEmbedding \longleftarrow [\targetEmbedding^1, \targetEmbedding^2, ..., \targetEmbedding^L]$\Comment{define target embedding}
      \State $D_{\set L}(\domainEmbedding, \targetEmbedding) = \sum_{l \in \set B} \set W (\domainEmbedding^l, \targetEmbedding^l) \quad \forall d \in D$\Comment{compute domain distances}
      \State $w^\targetdomain_d = \frac{1}{D_{\set L}(\domainEmbedding, \targetEmbedding)} \quad \forall d \in D$\Comment{compute domain similarities}
      \State $\domainPred \longleftarrow f(\targetSample|\domain) \quad \forall d \in D$\Comment{compute domain-specific predictions}
      \State $f(\targetSample) = \frac{\sum_{d \in \set D} w^\targetdomain_d \domainPred}{\sum_{d \in \set D} w^\targetdomain_d}$\Comment{compute final predictions}
  \EndFor
  \State$L(\theta;\macroBatch) = \sum_{(\targetSample,y_t) \in \macroBatch} \mathbb{XE}(f(\targetSample), y_t)$\Comment{compute cross-entropy loss}
  \State$\theta \longleftarrow \theta - \eta \cdot L(\theta;\macroBatch)$\Comment{update weights}
\end{algorithmic}
\end{algorithm*}

%
During every training step, first, the domain batches are propagated to update the corresponding domain embedding $\domainEmbedding$ \textit{(l:2-6)}.
Then, each individual sample $\targetSample$ is propagated using instance normalization to collect its instance statistics $(\mean_t^l,{\stddv 1_t^l}^2) \quad \forall l \in \set B$ \textit{(l:8)}.
Given the statistics we compute the target embedding $\targetEmbedding$ \textit{(l:9-10)} and the domain similarities $w^\targetdomain_d$ \textit{(l:12)}, as in~\textcolor{\refColor}{Sec. 3.3}.
Each sample is propagated under $\domainsetcardinality$ different domain assumptions (\ie through the corresponding domain-specific branches) \textit{(l:13)}.
The resulting domain-specific predictions are weighted according to~\textcolor{\refColor}{Eq. 11} to compute the final prediction \textit{(l:14)}.
Finally, the cross-entropy loss between the final predictions $f(\targetSample)$ and the corresponding ground truth $y_t$ is computed \textit{(l:15)} and back-propagated to update the weights $\theta$ of the model \textit{(l:16)}.
Applying this procedure during training encourages the creation of a batch normalization latent space.

\subsection{Training Settings} \label{ssec:training_settings}
Coherently with other works, we evaluate both the \alexnet{}~\cite{krizhevsky2012imagenet} and the more recent \resnet{}~\cite{he2016deep} architecture.
Before training each network, we initialize them with pre-trained weights on ImageNet and fine-tune the last fully-connected layer on the dataset of interest for $20$ epochs.
To train \alexnet{}~\cite{krizhevsky2012imagenet}, we use SGD as optimizer with momentum $0.95$ and L2 regularization on network weights with weight decay $5\times 10^{-5}$.
The initial learning rate is $10^{-3}$, exponentially decayed with decay rate $0.95$.
\resnet{} is trained with Adam~\cite{kingma2014adam} and weight decay $10^{-6}$. The initial learning rate is $10^{-4}$. 
Coherently with previous works~(\cite{carlucci2017just,carlucci2017autodial,mancini2018boosting}), we also compute gradients through the mean and standard deviation computation for the batch normalization layers.
All the input images are normalized according to the statistics computed on ImageNet. 
At training time, data augmentation is performed by first resizing the input image to $256 \time 256$, then randomly cropping to $224 \times 224$ for \resnet{} and $227 \times 227$ for \alexnet{}; finally, a random horizontal flip is performed.
Every training batch is composed of $16$ samples per domain for \resnet{} and $6$ for \alexnet{}.

All the models are implemented in Tensorflow 2.0~(\cite{abadi2015tensorflow}).
%
%
We initialize both \alexnet{} and \resnet{} using the publicly available Caffe weights pre-trained on ImageNet, after carefully converting them.\footnote{\resnet{} and \alexnet{} ImageNet weights available at \url{https://github.com/heuritech/convnets-keras} and \url{https://github.com/cvjena/cnn-models}.}

\subsection{Additional Results} \label{ssec:additional_results}
%
\input{tables/SOTA-PACS-Alexnet}
\input{tables/SOTA-Office-31-Resnet.tex}
\input{tables/SOTA-Office-Caltech-Resnet.tex}
%
We here provide additional results with the \resnet{}~\cite{he2016deep} architecture for the dataset \officeT{}~\cite{saenko2010adapting} and with the \alexnet{}~\cite{simon2016imagenet} architecture for \pacs{}~\cite{li2017deeper}.
In the original manuscript, we already provide results with \alexnet{} and \resnet{} respectively to compare against recently published works.
Moreover, we expand the experimental setting with the addition of the dataset \officecaltech{}~\cite{gong2012geodesic}, for which we present results with both \resnet{} and \alexnet{}.
%
%
%
\subsubsection{\pacs{}}
In \autoref{table:SOTA-PACS-Alexnet}, we extend the comparison on \pacs{} considering \alexnet{} to compare against a vast literature of published works relying on this older architecture.
Once again our proposal achieves absolute performance comparable to the state of the art even if starting from a weaker baseline. 
Indeed when comparing the relative gain in performance provided by our method (\gain{}), we are clearly outperforming any previously published solutions with an increase of $+6.33\%$, while the second best obtains $+4.88\%$. 
Once again, when considering \textit{Sketch} as unseen domain our method can boost the performance by a $+13\%$ absolute gain in accuracy over our baseline.
\subsubsection{\officeT{}}
In \autoref{table:SOTA-Office-31-Resnet}, we extend the comparison on \officeT{} considering \resnet{} as it is a good example of a modern architecture with native batch normalization layers.
The results confirms that our method is able to improve performances over \baseline{} across all three tests.
\subsubsection{\officecaltech{}}
%
%
%
%
%
In \autoref{table:SOTA-Office-Caltech-Resnet} we show additional results for \officecaltech{} using the \resnet{} architecture.
%
%
%
%
%
The same good property observed using Alexnet is confirmed also when considering ResNet as architecture in \autoref{table:SOTA-Office-Caltech-Resnet}, with a clear +5.5\% gain over \baseline{}.

\subsection{Ablation Study} \label{ssec:ablation}
We here provide additional ablation studies to better highlight different characteristics of our method with respect to the chosen batch size and the distance measures used in the latent space.
%
%
%
The experiments are conducted on the PACS dataset~\cite{li2017deeper}.
\subsubsection{Batch Size} \label{ssec:batch_size}
\input{tables/batchsize.tex}

We study the impact of different batch sizes on the performance of our method in \autoref{table:batchsize}.
As expected and already documented in several recent works leveraging batch normalization layers~\cite{bjorck2018understanding,wu2018group}, the larger the batch size is the better the generalization capability.
In particular for our method the bigger is the batch size used at training time, the better are the approximation of the true population statistics (\ie, the better are the domain embeddings $\domainEmbedding{}$).
This translates in better final performance as detailed in \autoref{table:batchsize} where we can observe an increment of $+2.9$ \textit{Average} accuracy between using batch size 16 and 64.

\subsubsection{Method Components}
\label{ssec:method_components}
In the main paper we measured the contribution in achieving the final performance of the different components of our methods. 
The proposed setting leveraged the \pacs{} dataset and the \resnet{} architecture.
We here consider ablation experiments on the \pacs{} dataset using the \alexnet{} architecture and report the results in \textcolor{\refColor}{Tab.3}, 
comparing again with the \baseline{} baseline.
On row (a) we show the performance gained by using separate batchnorm statistics for the different train domains and using the projection and weighting strategy described in~\textcolor{\refColor}{Sec. 3.3}; row (b) extends the method above by using the \textit{distance weighting at training time} (DT) as described in~\textcolor{\refColor}{Sec. 3.4}; finally, row (c) includes a warm-up phase in the training of the model to make population statistics converge to stable values before starting the distance training.
By comparing the average accuracy (\averagePerfomance{}) across the four possible target sets, it is clear how every component contributes to an increase in performance with respect to the baseline.
%
\input{tables/method_alexnet}

\subsubsection{Normalization Strategies}
\label{ssec:normalization_strategies}
%
\input{tables/normalization_strategy}
\methodName{} can be interpreted as a peculiar normalization technique to achieve better generalization.
We hence provide a quantitative comparison of our method against other popular normalization strategies:
(a) InstanceNorm~\cite{li2018adaptive};
(b) BatchNorm~\cite{ioffe2015batch};
(c) Freeze BatchNorm~\cite{ioffe2015batch} (\ie, keeping the population statistics as the one computed after the imagenet pre-training);
(d) \methodName \textit{(Ours)}.

Results for this comparison are shown in \autoref{table:normalization_strategies}.
Freezing batch normalization statistics (c) to those accumulated on ImageNet provides better generalization than fine-tuning population statistics on the training datasets (b), which might lead to overfitting and is equivalent to the baseline \baseline{}.
This is coherent with what highlighted in~\cite{seo2019learning},
Among the analysed normalization techniques, InstanceNorm (a) achieves the poorest results.
Our proposal (d) instead, by combining instance and batch normalization properties in a principled way, achieves the best results.
We can indeed notice how \methodName{} outperforms by a large margin all other normalization strategies, both overall ($+3.5\%$ over Freeze BatchNorm) and on any specific domain.

\subsection{Latent Space Validation}
\label{ssec:latentSpaceExperimental}
We now want to investigate how well we are able to collect domain specific attributes of samples by  projecting them to the batchnorm latent space.
We trained \resnet{} until convergence without distance training and warm-up on the \pacs{} dataset considering \textit{Photo} or \textit{Sketch} as unseen domains.
Once trained, we forward every training sample through the network and compute its instance statistics to project it to the batchnorm latent space.
After the projection we measure the distance from every domain embedding: if the closest domain matches the real domain, then the latent space effectively represents membership to a certain domain.
In \autoref{table:weights} we report the average value of the reciprocal of the distance for every training sample with respect to the centroid of the three training domains.
The higher values on the diagonal confirm our intuition that the batchnorm latent space can be used to implicitly encode domain attributes.
%
\input{tables/weights.tex}

Furthermore, we investigate the relationship between measured distances and the prediction accuracy on the same ResNet-18 trained on PACS without DT.
%
For each test sample we measure the prediction accuracy obtained using only the predictions from either the closest domain branch, the second closest or the third (i.e., the farthest away).
We run the test for all 4 possible unseen domains following the leave-one-domain-out protocol and report in \autoref{table:closest_domain_accuracy} the average accuracy.
%
The results show a clear correlation between distances and accuracy, as trusting the ``closest" domain branch clearly results in a higher accuracy than the others.
%
\input{tables/closest_domain_accuracy.tex}

\subsection{Domain Discovery Net}
%
\input{tables/domain_accuracy.tex}
%
In~\textcolor{\refColor}{Sec 4.3}, we compared the performance of~\methodName{} with~\discoveryName{}. 
For this purpose, we follow~\cite{mancini2018robust} and implement a domain discovery network (\discoveryName{}) that takes as input the activations after the first convolutional block and directly outputs the probability for the input sample to belong to each one of the training domains.
This probability distribution is used to weigh the domain-specific predictions of our lightweight ensemble.
Analogously to \cite{mancini2018robust} we implemented \discoveryName{} as a lateral branch to our lightweight ensemble that is composed of a global pooling layer, followed by a ReLU non linearity, a fully-connected layer and a softmax activation.
\discoveryName{} is trained in an end-to-end fashion together with the main classifier.
We considered two options: (i) training \discoveryName{} applying only a cross-entropy loss on the image classification logits with respect to the input categories; (ii) training \discoveryName{} directly supervising the classification of samples in the correct domain using domain labels.

In~\autoref{table:domain_accuracy} we compare the domain classification accuracy (\averageDomain) of~\methodName{} and that of~\discoveryName{} with or without applying a cross entropy loss over the domain labels across four tests considering different unseen domains on \pacs{}.
When cross-entropy is not applied on domain logits,~\methodName{} largely outperforms~\discoveryName{}.
The $33\%$ \averageDomain for~\discoveryName{} without cross-entropy on domain logits denotes that the discovery network learns to disregard the multidomain BN layer, always predicting the same domain class and thus leveraging only one branch of the multidomain BN layer.
However, when cross-entropy is applied also on domain logits, the domain classification branch of~\discoveryName{} can adapt its parameters to predict well the domain classes (\averageDomain).
This, however, comes at the cost of a remarkable drop in image classification accuracy (\averageClass) that can may be partially explained by \discoveryName{} overfitting more to the training data and being less able to generalize to the test one. 
\methodName{} instead provides a meaningful domain representation even without cross-entropy on domain logits, largely outperforming the image classification accuracy of~\discoveryName{}.
Nevertheless, since our representation is not parametric, we cannot witness a visible increase in~\averageDomain when applying a cross-entropy loss also on the domain membership assigned through our representation.

The advantages of~\methodName{} over~\discoveryName{} are clear, since~\methodName{} leverages all the activations throughout the network to get an estimate of the domain membership, while~\discoveryName{} must rely only on the activations of the first layer due to the fixed input size of the domain classification branch.
Moreover, our method allows a parameter-free domain representation, while~\discoveryName{} relies on a lateral branch to the main network.
Finally,~\methodName{} allows to map samples in a latent space where distances from domain embeddings are computed, while~\discoveryName{} can only output the distance of the input sample from the training domains. 

%% file: tables/SOTA-PACS-Alexnet.tex
%
\setlength{\tabcolsep}{4pt}
\begin{table*}[t]
\begin{center}
\caption{State-of-the-art comparison on PACS with~\alexnet{}.}
\label{table:SOTA-PACS-Alexnet}
\begin{tabular}{M{5.0cm}|M{1.1cm} M{1.1cm} M{1.1cm} M{1.1cm}|M{1.3cm} M{1.0cm} M{1.0cm}}
\hline\noalign{\smallskip}
Method & Art & Cartoon & Photo & Sketch & \averageDeepAll{} & \averagePerfomance{} & $\Delta \%$ \\
\noalign{\smallskip}
\hline
\noalign{\smallskip}
DICA -~\cite{muandet2013domain} & 64.6 & 64.5 & 91.8 & 51.1 & 68.7 & 68.0 & -1.02 \\
D-MTAE -~\cite{ghifary2015domain} &  60.3 & 58.7 & 91.1 & 47.9 & 68.7 & 64.5 & -6.11 \\
DSN -~\cite{bousmalis2016domain} &  61.1 & 66.5 & 83.3 & 58.6 & 68.7 & 67.4 & -1.89 \\
TF-CNN -~\cite{li2017deeper} &  62.9 & 67.0 & 89.5 & 57.5 & 67.1 & 69.2 & +3.13 \\
CIDDG -~\cite{li2018deep} &  62.7 & 69.7 & 78.7 & 64.5 & 71.7 & 68.9 & -3.91 \\
Fusion -~\cite{mancini2018best} &  64.1 & 66.8 & 90.2 & 60.1 & 67.1 & 70.3 & +4.77 \\
CrossGrad -~\cite{shankar2018generalizing} &  64.1 & 66.8 & 90.2 & 60.1 & 68.7 & 70.3 & +2.33\\
MetaReg -~\cite{balaji2018metareg} &  69.8 & 70.4 & 91.1 & 59.3 & 69.3 & 72.6 & +4.76 \\
MLDG -~\cite{li2018learning} &  66.2 & 66.9 & 88.0 & 59.0 & 67.2 & 70.0 & +4.17 \\
Epi-FCD -~\cite{li2019episodic} &  64.7 & 72.3 & 86.1 & 65.0 & 68.7 & 72.0 & +4.80 \\
JiGen -~\cite{carlucci2019domain} &  67.6 & 71.7 & 89.0 & 65.2 & 71.5 & 73.4 & +2.66 \\
MASF -~\cite{dou2019domain} &  \textbf{70.4} & \textbf{72.5} & \textbf{90.7} & \textbf{67.3} & 71.7 & \textbf{75.2} & +4.88 \\
\noalign{\smallskip}
\hline
\noalign{\smallskip}
DeepAll & 64.4 & 65.4 & 88.0 & 53.8 & - & 67.9 & - \\
\methodName{} (\textit{Ours}) &  66.7 & 65.7 & 89.5 & 66.8 & 67.9 & 72.2 & \textbf{+6.33} \\
\hline
\end{tabular}
\end{center}
\end{table*}
\setlength{\tabcolsep}{1.4pt}
%

%% file: tables/SOTA-Office-31-Resnet.tex
%
\setlength{\tabcolsep}{4pt}
\begin{table*}[h!]
\begin{center}
\caption{State-of-the-art comparison on Office-31 with~\resnet{}.}
\label{table:SOTA-Office-31-Resnet}
\begin{tabular}{M{2.0cm}|M{1.2cm} M{1.2cm} M{1.2cm}|M{1.3cm} M{1.2cm} M{1.2cm}}
\hline\noalign{\smallskip}
Method & Amazon & Dslr & Webcam & \averageDeepAll{} & \averagePerfomance{} & $\Delta \%$ \\
\noalign{\smallskip}
\hline
\noalign{\smallskip}
DeepAll & 55.1 & 99.0 & 92.6 & - & 82.2 & - \\
\methodName{} (\textit{Ours}) & \textbf{55.5} & \textbf{99.3} & \textbf{95.4} & 82.2 & \textbf{83.4} & +1.42 \\
\hline
\end{tabular}
\end{center}
\end{table*}
\setlength{\tabcolsep}{1.4pt}
%

%% file: tables/SOTA-Office-Caltech-Resnet.tex
%
\setlength{\tabcolsep}{4pt}
\begin{table*}[h!]
\begin{center}
\caption{State-of-the-art comparison on Office-Caltech with~\resnet{}.}
\label{table:SOTA-Office-Caltech-Resnet}
\begin{tabular}{M{2.2cm}|M{1.1cm} M{1.1cm} M{1.1cm} M{1.2cm}|M{1.3cm} M{1.0cm} M{1.0cm}}
\hline\noalign{\smallskip}
Method & Amazon & Caltech & Dslr,   & Amazon, & \averageDeepAll{} & \averagePerfomance{} & $\Delta \%$ \\
       &        &         &  Webcam & Caltech &         &         &   \\
\noalign{\smallskip}
\hline
\noalign{\smallskip}
DeepAll & 92.7 & 83.1 & 85.3 & 80.7 & - & 85.5 & - \\
\methodName{} (\textit{Ours}) &  \textbf{92.9} & \textbf{87.4} & \textbf{93.0} & \textbf{87.3} & 85.5 & \textbf{90.2} & +5.50 \\
\hline
\end{tabular}
\end{center}
\end{table*}
\setlength{\tabcolsep}{1.4pt}
%

%% file: tables/batchsize.tex
\setlength{\tabcolsep}{4pt}
\begin{table*} [h!]
\begin{center}
\caption{Comparison of different batch sizes (per domain) on PACS with Alexnet.}
\label{table:batchsize}
\begin{tabular}{M{1.8cm}|M{1.6cm}|M{1.2cm}M{1.2cm}M{1.2cm}M{1.2cm}|M{1.2cm}}
\hline\noalign{\smallskip}
Method & Batch Size & Art & Cartoon & Photo & Sketch & Average \\
\noalign{\smallskip}
\hline
\noalign{\smallskip}
DeepAll & - & 64.4 & 65.4 & 88.0 & 53.8 & 67.9 \\
\methodName{} (\textit{Ours}) & 16 & 65.7 & 66.5 & 88.9 & 56.0 & 69.3 \\
\methodName{} (\textit{Ours}) & 32 & 67.9 & 65.7 & 89.4 & 62.3 & 71.3 \\
\methodName{} (\textit{Ours}) & 64 & 66.7 & 65.7 & 89.5 & 66.8 & 72.2 \\
\hline
\end{tabular}
\end{center}
\end{table*}
\setlength{\tabcolsep}{1.4pt}
%

%% file: tables/method_alexnet.tex
\setlength{\tabcolsep}{4pt}
\begin{table*}[t]
\begin{center}
\caption{Comparison of different variants of our method on PACS with Alexnet.}
\label{table:method_alexnet}
\begin{tabular}{M{1.5cm}|M{0.8cm}M{1.4cm}|M{1.2cm}M{1.2cm}M{1.2cm}M{1.2cm}|M{1.2cm}}
\hline\noalign{\smallskip}
Method & DT & Warm-up & Art & Cartoon & Photo & Sketch & Avg. \\
\noalign{\smallskip}
\hline
\noalign{\smallskip}
DeepAll & - & - & 64.4 & 65.4 & 88.0 & 53.8 & 67.9 \\
\midrule
(a) \methodName{} & \xmark & \xmark & 64.4 & 65.9 & \textbf{89.6} & 54.2 & 68.53 \\
(b) \methodName{} & \cmark & \xmark & 63.7 & \textbf{67.9} & 84.6 & 66.6 & 70.7 \\
(c) \methodName{} & \cmark & \cmark & \textbf{66.7} & 65.7 & 89.5 & \textbf{66.8} & \textbf{72.2} \\
\hline
\end{tabular}
\end{center}
\end{table*}
\setlength{\tabcolsep}{1.4pt}
%

%% file: tables/normalization_strategy.tex

\setlength{\tabcolsep}{4pt}
\begin{table*} [h!]
\begin{center}
\caption{Comparison of different normalization strategies on PACS with \resnet{}.}
\label{table:normalization_strategies}
\begin{tabular}{M{4.0cm}|M{1.2cm}M{1.2cm}M{1.2cm}M{1.2cm}|M{1.2cm}}
\hline\noalign{\smallskip}
Method & Art & Cartoon & Photo & Sketch & Average \\
\noalign{\smallskip}
\hline
\noalign{\smallskip}
(a) InstanceNorm & 62.6 & 72.7 & 79.7 & 71.7 & 71.7 \\
(b) BatchNorm & 75.8 & 73.0 & 94.4 & 70.9 & 78.5 \\
(c) Freeze BatchNorm & 75.0 & 76.8 & 92.6 & 73.9 & 79.6 \\
(d) BNE (\textit{Ours}) & \textbf{78.8} & \textbf{78.9} & \textbf{94.8} & \textbf{79.7 } & \textbf{83.1} \\
\hline
\end{tabular}
\end{center}
\vspace{-0.3cm}
\end{table*}
\setlength{\tabcolsep}{1.4pt}

%% file: tables/weights.tex
\setlength{\tabcolsep}{4pt}
\begin{table}[t]
\begin{center}
\caption{Analysis of the average similarity value as domain classification metrics with \resnet{} on \pacs{} without distance training. Classified domain is in bold. }
\label{table:weights}
\subfloat[\textit{Photo} unseen.]{\begin{tabular}{M{1.2cm}|M{1.1cm}M{1.1cm}M{1.1cm}}
\hline\noalign{\smallskip}
Source & Art & Cartoon & Sketch \\
\noalign{\smallskip}
\hline
\noalign{\smallskip}
Art & \textbf{8.24} & 5.35 & 4.21 \\
\noalign{\smallskip}
\hline
\noalign{\smallskip}
Cartoon & 6.58 & \textbf{7.02} & 5.97 \\
\noalign{\smallskip}
\hline
\noalign{\smallskip}
Sketch & 3.94 & 4.56 & \textbf{10.19} \\
\noalign{\smallskip}
\hline
\end{tabular}}
\quad
\subfloat[\textit{Sketch} unseen.]{\begin{tabular}{M{1.2cm}|M{1.1cm}M{1.1cm}M{1.1cm}}
\hline\noalign{\smallskip}
Source & Art & Cartoon & Photo \\
\noalign{\smallskip}
\hline
\noalign{\smallskip}
Art & \textbf{1.18} & 0.70 & 1.15 \\
\noalign{\smallskip}
\hline
\noalign{\smallskip}
Cartoon & 0.94 & \textbf{1.02} & 0.90 \\
\noalign{\smallskip}
\hline
\noalign{\smallskip}
Photo & 1.19 & 0.70 & \textbf{1.25} \\
\noalign{\smallskip}
\hline
\end{tabular}}
\end{center}
\vspace{-0.5cm}
\end{table}
\setlength{\tabcolsep}{1.4pt}
%

%% file: tables/closest_domain_accuracy.tex
%
\setlength{\tabcolsep}{8pt}
\begin{table*}[h]
\begin{center}
\caption{Analysis of the classification accuracy considering predictions from the closest, second-closest (second) and third-closest (third) domain branches to target sample. We also report the average accuracy (Avg.) over all leave-one-domain-out tests.}
\label{table:closest_domain_accuracy}
\begin{tabular}{c|c c c c|c }
\hline\noalign{\smallskip}
Method & Art Painting & Cartoon & Photo & Sketch & Avg. \\
\noalign{\smallskip}
\hline
\noalign{\smallskip}
Closest & \textbf{68.02} & \textbf{63.10} & \textbf{92.75} & \textbf{71.09} & \textbf{72.20} \\
Second & 65.14 & 60.75 & 86.83 & 57.11 & 64.58 \\
Third & 47.95 & 60.66 & 67.31 & 57.90 & 58.08 \\
\hline
\end{tabular}
\end{center}
\end{table*}
%

%% file: tables/domain_accuracy.tex
\setlength{\tabcolsep}{4pt}
\begin{table*} [h!]
\begin{center}
\caption{Comparison \methodName{} and \discoveryName{} with or without cross-entropy loss applied also on domain logits. We report the domain classification accuracy for different runs with different unseen domains, the average domain classification accuracy (Avg. Domain) and the average image classification accuracy (Avg. Class).}
\label{table:domain_accuracy}
\begin{tabular}{M{1.8cm}|M{0.6cm}|M{1.2cm}M{1.2cm}M{1.2cm}M{1.2cm}|M{2.0cm}|M{1.9cm}}
\hline\noalign{\smallskip}
Method & XE & Art & Cartoon & Photo & Sketch & \averageDomain & \averageClass \\
\noalign{\smallskip}
\hline
\noalign{\smallskip}
\methodName{} & \xmark & \textbf{71.2} & \textbf{82.2} & \textbf{75.0} & \textbf{60.1} & \textbf{72.1} & \textbf{83.1} \\
\discoveryName{} & \xmark & 33.3 & 33.3 & 33.3 & 33.3 & 33.3 & 79.1 \\
\noalign{\smallskip}
\hline
\noalign{\smallskip}
\methodName{} & \cmark & 75.9 & 82.7 & 74.5 & 58.7 & 73.0 & \textbf{80.8} \\
\discoveryName{} & \cmark & \textbf{96.0} & \textbf{87.8} & \textbf{62.7} & \textbf{84.3} & \textbf{82.7} & 74.9 \\
\hline
\end{tabular}
\end{center}
\end{table*}
\setlength{\tabcolsep}{1.4pt}
%